%% file: ijcai24.tex
\documentclass{article}
\usepackage{styles/ijcai24}

\usepackage{times}
\usepackage{soul}
\usepackage{url}
\usepackage[hidelinks]{hyperref}
\usepackage[utf8]{inputenc}
\usepackage[small]{caption}
\usepackage{graphicx}
\usepackage{amsmath}
\usepackage{amsthm}
\usepackage{booktabs}
\usepackage{algorithm}
\usepackage{algorithmic}
\usepackage[switch]{lineno}
\usepackage{subcaption}
\usepackage{multirow}
\usepackage{amssymb}
\usepackage{pifont}
\usepackage{tabularx}
\usepackage{mdframed}
\usepackage{xcolor}

\title{Scene-Adaptive Person Search via Bilateral Modulations}

\author{
Yimin Jiang$^1$
\and
Huibing Wang$^1$\thanks{Corresponding Author}\and
Jinjia Peng$^2$\and
Xianping Fu$^1$\And
Yang Wang$^3$\\
\affiliations
$^1$School of Information Science and Technology, Dalian Maritime University, Dalian, China\\
$^2$School of Cyber Security and Computer, Hebei University, Baoding, China\\
$^3$Key Laboratory of Knowledge Engineering with Big Data, Ministry of Education, Hefei University of Technology, Hefei, China\\
\emails
\{yimin\_jiang, huibing.wang, fxp\}@dlmu.edu.cn,
pengjinjia@hbu.edu.cn,
yangwang@hfut.edu.cn
}

\graphicspath{{figures/}}
\newcommand{\tstretch}{1.}  

\begin{document}

\maketitle

\input{sections/0_abstract}
\input{sections/1_introduction}

\input{sections/3_method}
\input{sections/4_experiments}
\input{sections/5_conclusion}

\appendix

\bibliographystyle{styles/named}
\bibliography{sections/ijcai24}

\end{document}

%% file: sections/0_abstract.tex
\begin{abstract}
    \label{sec:abstract}
    Person search aims to localize specific a target person from a gallery set of images with various scenes. As the scene of moving pedestrian changes, the captured person image inevitably bring in lots of background noise and foreground noise on the person feature, which are completely unrelated to the person identity, leading to severe performance degeneration. To address this issue, we present a Scene-Adaptive Person Search (SEAS) model by introducing bilateral modulations to simultaneously eliminate scene noise and maintain a consistent person representation to adapt to various scenes. In SEAS, a Background Modulation Network (BMN) is designed to encode the feature extracted from the detected bounding box into a multi-granularity embedding, which reduces the input of background noise from multiple levels with norm-aware. Additionally, to mitigate the effect of foreground noise on the person feature, SEAS introduces a Foreground Modulation Network (FMN) to compute the clutter reduction offset for the person embedding based on the feature map of the scene image. By bilateral modulations on both background and foreground within an end-to-end manner, SEAS obtains consistent feature representations without scene noise. SEAS can achieve state-of-the-art (SOTA) performance on two benchmark datasets, CUHK-SYSU with 97.1\% mAP and PRW with 60.5\% mAP. The code is available at \href{https://github.com/whbdmu/SEAS}{\footnotesize\textcolor{magenta}{\textit{\texttt{https://github.com/whbdmu/SEAS}}}}.
\end{abstract}

%% file: sections/1_introduction.tex
\section{Introduction}
\label{sec:introduction}

\begin{figure}[t]
    \centering
    \includegraphics[scale=0.57]{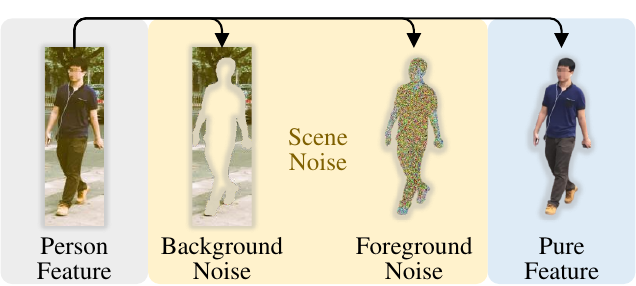}
    \caption{Composition of the person feature. The person feature consists of scene noise and pure person feature, while the scene noise can be divided into background noise, which comes from the residual background in the detected bounding box, and foreground noise, which is caused by the influence of lighting conditions, visibility, and so on.}
    \label{fig:noise}
\end{figure}

\mdfdefinestyle{comparison}{
    linewidth=.6pt,
    innertopmargin=0.4em,
    innerbottommargin=0.4em,
    innerleftmargin=0.em,
    innerrightmargin=0.em,
    leftmargin=0.em,
    rightmargin=0.em,
}

\begin{figure}[h]
    \centering
    \begin{subfigure}{0.186\linewidth}
        \centering
        \begin{mdframed}[style=comparison]
            \centering
            \includegraphics[scale=0.57]{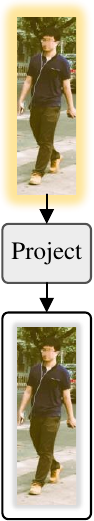}
        \end{mdframed}
        \caption{\scriptsize Projection}
        \label{fig:a}
    \end{subfigure}
    \hspace{-6pt}
    \begin{subfigure}{0.407\linewidth}
        \centering
        \begin{mdframed}[style=comparison]
            \centering
            \includegraphics[scale=0.57]{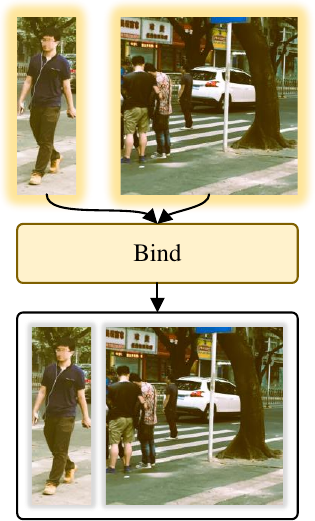}
        \end{mdframed}
        \caption{\scriptsize Binding}
        \label{fig:b}
    \end{subfigure}
    \hspace{-6pt}
    \begin{subfigure}{0.407\linewidth}
        \centering
        \begin{mdframed}[style=comparison]
            \centering
            \includegraphics[scale=0.57]{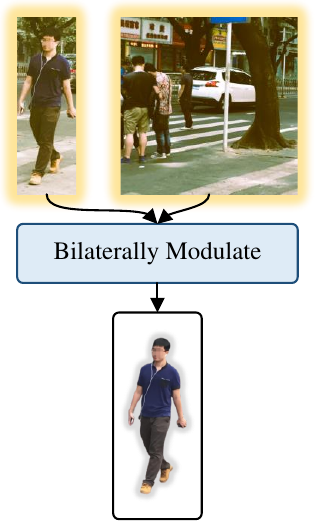}
        \end{mdframed}
        \caption{\scriptsize Ours}
        \label{fig:c}
    \end{subfigure}
    \caption{Comparison of three person search strategies. (a) Using the projected person feature to search is the initial method, but its cross-scene ability is unsatisfactory due to the neglect of scene noise. (b) The binding strategy is to bind scene features to character features. It has excellent person retrieval when each person is in a fixed scene, but changing the scene causes the retrieval to deteriorate. (c) Our strategy is to leverage scene features to eliminate scene noises from person features, achieving adaptation to diverse scenes.}
    \label{fig:comparison}
\end{figure}

Person search aims to localize and identify the query person from a gallery of the whole scene images, which can be taken as a joint task of person detection and re-identification (re-id) \cite{peng2021unsupervised,wang2022progressive} in computer vision \cite{wang2024unpacking}. Because person search can directly derive person identities from real-world scenarios, it has attracted wide attention from various applications. However, with the moving pedestrian, the varied scene not only bring in background noise but also cause a negative impact on person features, which inevitably leads to severe performance degradation, as shown in Figure \ref{fig:noise}. Therefore, identifying the same individual across varied scenes is much more challenging than in a fixed setting and cries out for reliable solutions.

Nowadays, to deal with this problem, a series of works have been developed and can be divided into two categories. The first category of methods performs person search directly employing the projection of person feature extracted from the detected bounding box, as shown in Figure \ref{fig:a}, which is the mainstream practice in this field. Xiao et al. design the first end-to-end network, where detection and feature embedding share a head network \cite{xiao2017joint}. SeqNet \cite{li2021sequential} sequentially tackles detection and re-id in an end-to-end model, and assigns an exclusive head network to feature embedding. Even though some of them can achieve satisfactory performance, they do not realize and eliminate the additional noise from the scene, resulting in severe performance drops facing changing scenes. The second category acknowledges the auxiliary role of scene information in person search and explores the way to take it into account. GLCNet \cite{zheng2021global} is proposed to embed scene information into person representation to improve the discrimination of different persons from different scenes. GFN \cite{jaffe2023gallery} calculates the similarity of the scenes where persons are, and then filters out images with low scene similarity. These methods essentially adopt a strategy to bind the scene and the person, as shown in Figure \ref{fig:b}. Assuming a fixed scene for each person, this strategy can significantly improve the performance of person search. However, this assumption always fails in real-world applications, and the binding strategy has the opposite effect, because the changing scene of the same person will make it impossible to have a consistent identity representation. Hence, it is imperative to acknowledge the influence of scene on person search and exploit scene information judiciously. Notably, a scene can directly lead to two effects on person search, including residual background from the detected bounding box and foreground noise caused by lighting conditions, shadow changes, etc.. Therefore, how to remove these effects from the scene to keep a consistent person representation across different scenes for a same person is crucial but challenging.

To relieve this dilemma, we present a novel Scene-Adaptive Person Search (SEAS) method that adopts bilateral modulations to exploit the scene feature to eliminate the scene noise presented in the person feature, thus obtaining a pure person feature for searching to adapt to diverse scenes, as shown in Figure \ref{fig:c}. To eliminate the residual background in the detected bounding box, SEAS proposes a Background Modulation Network (BMN) that incorporates our designed Multi-Granularity Embedding (MGE) to encode the feature embedding that integrate various granularities of person information to reduce the background noise at multiple levels instead of only at the global level. Meanwhile, BMN proposes a Background Noise Reduction (BNR) loss to improve the network's ability to specifically shield against background noise. To eliminate the effect of scene on the foreground of person, SEAS proposes a Foreground Modulation Network (FMN), which employs the scene feature to correct the deviation of person feature caused by the foreground noise. Specifically, FMN constructs a scene noise extractor and a person feature denoiser. The denoiser applies a cross-attention mechanism between the person feature embedding and the scene noise map derived by the extractor to obtain a offset which is added to the feature embedding to counteract the foreground noise. By applying bilateral modulations to person features, SEAS can adapt various scenes to achieve SOTA performance for person search.

In summery, our paper makes three contributions:

\begin{itemize}
    \item This paper proposes a Scene-Adaptive Person Search (SEAS) framework that achieves scene adaptation by maintaining highly similar feature representations of the same person in different scenes.
    \item A Background Modulation Network (BMN) is designed to generate multi-granularity feature embeddings based on person feature and to suppress background noise at multiple levels.
    \item To the best of our knowledge, our Foreground Modulation Network (FMN) is the first to use scene information to eliminate foreground noise.
\end{itemize}

%% file: sections/3_method.tex
\section{Method}
\label{sec:method}

In this section, we introduce the proposed Scene-Adaptive Person Search (SEAS) framework. Firstly, we give an overview of the network architecture. Secondly, the proposed Background Modulation Network (BMN) is elaborated, which aims to filter out the background noise and encode the person feature at multiple granularities. Thirdly, the designed Foreground Modulation Network (FMN) is presented, which aims to eliminating the foreground noise. Finally, we present a re-id loss named Bidirectional Online Instance Matching (BOIM).

\subsection{Framework overview}

\begin{figure*}
    \centering
    \includegraphics[scale=0.57]{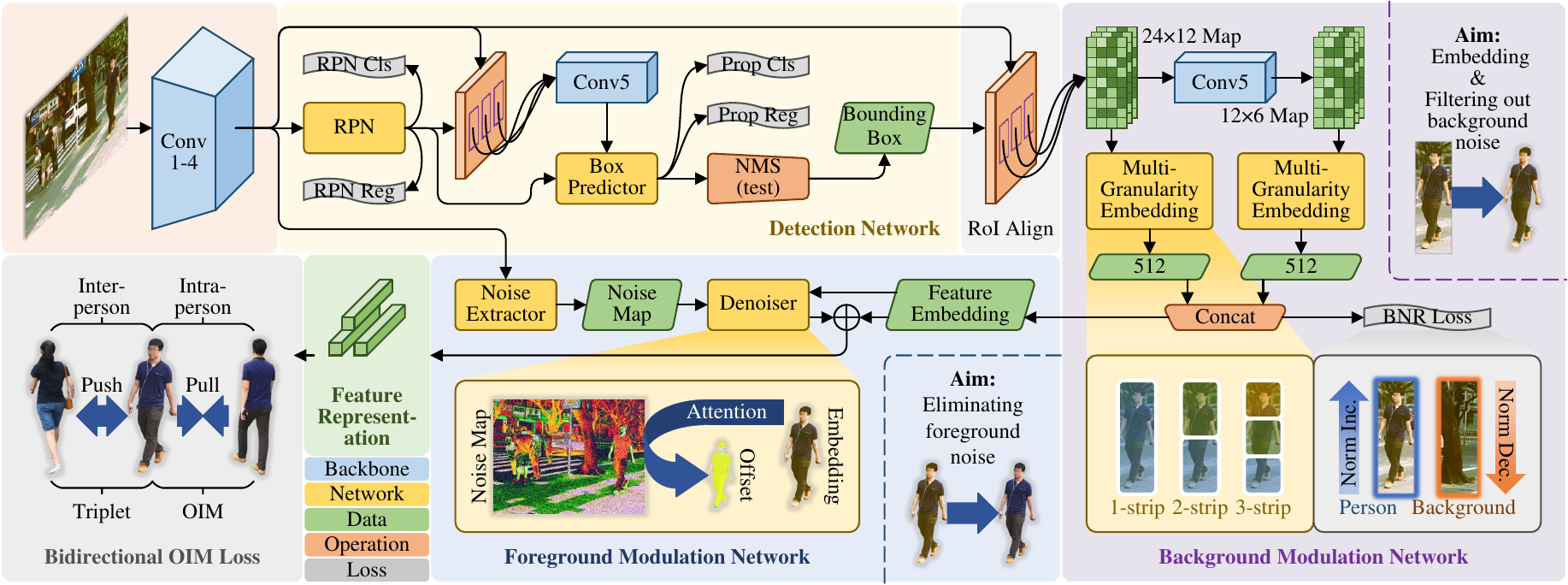}
    \caption{Architecture of the SEAS framework. The lower left corner of this figure is marked with the meaning indicated by the color of component. This figure can be divided into two rows, in a clockwise direction, starting at the top left and ending at the bottom left. Wrapped in a solid rounded box is the schematic of the component; wrapped in a dashed rounded box is the aim for the network.}
    \label{fig:main_arch}
\end{figure*}

Our person search framework is an end-to-end architecture based on SeqNet \cite{li2021sequential}. Differently, we adhere to the principle of ``re-identification first'' and use our novel method to obtain robust feature representations to adapt to various scenes. Therefore, we remove all detection-related components from the re-id head network and carefully design bilateral modulations, containing a background modulation network and a foreground modulation network, as a new re-id head network, as shown in Figure \ref{fig:main_arch}.

SEAS consists of four components: a backbone, a detection network, a background modulation network, and a foreground modulation network. We use the first four convolution blocks of ConvNeXt \cite{liu2022convnet} as the backbone to extract a scene feature map from an entire scene image. The feature map is then fed into the detection network, where the region proposals generated by RPN \cite{ren2015faster} are classified and regressed by the box predictor and output as detected bounding boxes. RoI-Align is exploited to pool a person feature map from the scene feature map for each detected bounding box. The person feature map is encoded by the BMN into 1,024-dimensional feature embedding at multiple granularities, and its accompanying background noise is suppressed. Finally, the FMN uses the scene feature map to calculate an offset for the feature embedding, thereby removing the foreground noise contained in the person feature.

During the training phase, the feature representation is supervised by our proposed bidirectional online instance matching loss, which is a combination of OIM \cite{xiao2017joint} loss and our designed triplet loss. Among them, the triplet and OIM losses respectively focus on encouraging our model to learn inter-person differences and intra-person similarities.

\subsection{Background Modulation Network}

In BMN, we make use of two levels of a person's feature maps for embedding, so that the representation has different scales of the feature. Specifically, the BMN performs MGEs on the 24×12 feature map output by the RoI-Align \cite{girshick2015fast} and the 12×6 feature map down-sampled by the Conv5 which is the fifth convolution block of ConvNeXt, respectively, and concatenates these two outputs as the feature embedding, as shown in Figure \ref{fig:main_arch}. For the feature embedding, under the supervision of the proposed BNR loss, there is almost no background noise.

\begin{figure}[h]
    \centering
    \includegraphics[scale=0.57]{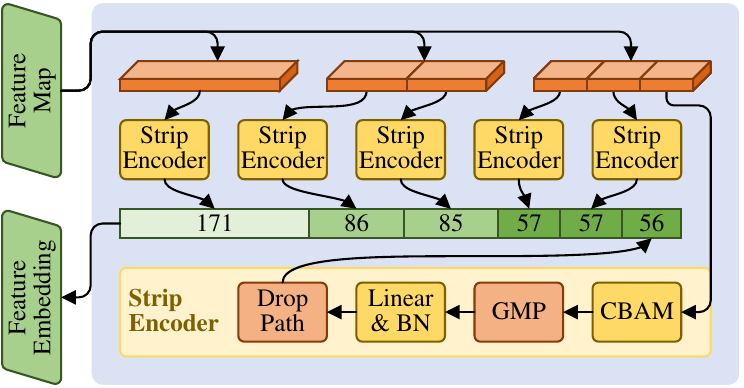}
    \caption{Details of our multi-granularity embedding.}
    \label{fig:mge}
\end{figure}

\textbf{Multi-Granularity Embedding}. The global feature embedding aims to capture the most salient appearance symbol of a person as the representation of its identity. However, in global feature learning, some detailed clues that can contribute to distinguishing the person is easily discarded. So we utilize the proposed MGE to combine global and local features of the person to obtain a accurate feature representation. MGE consists of three branches. The first branch learns the global feature of the person, while the second and third branches evenly divide the person feature map into 2 and 3 horizontal strips, respectively, and independently learn the local feature on each strip. We carefully design a Strip Encoder to handle the strip, and the global feature is also treated as one strip, as shown in Figure \ref{fig:mge}. The encoder comprises five components: a CBAM \cite{woo2018cbam}, a global max pooling, a linear transformation, a batch normalization \cite{ioffe2015batch} and a drop path \cite{larsson2016fractalnet}. The CBAM is a channel and spatial attention module that is combined with the global max pooling to filter out background noise under the supervision of BNR loss. The batch normalization normalizes the value to around 0 to prevent it from clustering in positive or negative intervals, thereby improving the discrimination of the embedding. Through the Drop Path, a robust embedding is obtained, as it hinders the embedding from excessively relying on a specific strip.

\textbf{Background Noise Reduction Loss.} The detected bounding box contains some residual background in addition to the person, which is referred to as background noise. While the attention characteristics of the network can filter out the background noise, this capability is weak because it is learned as an ancillary task for extracting the person feature. Therefore, inspired by NAE \cite{chen2020norm}, a Background Noise Reduction (BNR) loss is proposed to enhance the network's ability to shield against the background noise specifically. BNR loss calculates the L$_2$-norm of feature embedding for each detected box and uses its corresponding label (person or background) to supervise it. To make the norm meet the condition of binary cross entropy, we use the following linear mapping to compress its size from $\left[0,\infty \right)$ to the range of $\left(0, 1\right)$.
\begin{equation}
    q = \textrm{sigmoid}\left[\textrm{BN}\left( \left\| \xi \right\|_2 \right) \right]
    \label{equ:bnr-prob}
\end{equation}
where $\left\| \xi \right\|_2$ is the L$_2$-norm of the feature embedding $\xi$. The loss calculation follows:
\begin{equation}
    \mathcal{L}_{bnr} = -\left[ y\log\left( q \right) + \left( 1-y \right) \log \left( 1-q \right) \right]
    \label{equ:bnr}
\end{equation}
where $y$ is the label (either $0$ or $1$) indicating whether its corresponding embedding is classified as a background or a person. This method amplifies the norm for the person and diminishes it for the background, thereby significantly reducing the sensitivity of the network to backgrounds.

\subsection{Foreground Modulation Network}

\begin{figure}[t]
    \centering
    \includegraphics[scale=0.57]{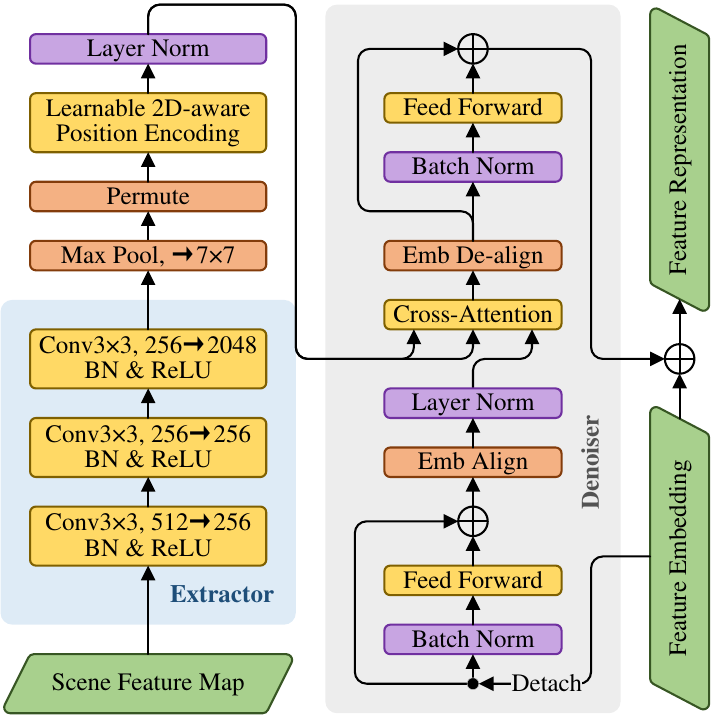}
    \caption{Details of our foreground modulation network.}
    \label{fig:fmn}
\end{figure}

The expression of foreground noise is salient and rich in the scene image, but it is difficult to perceive in the person feature due to the localized nature of person image in the scene image. Therefore, we utilize the scene feature to correct the deviation of person feature caused by the foreground noise. In FMN, as shown in Figure \ref{fig:fmn}, a noise extractor is designed to extract a noise map from the scene feature map, and a feature denoiser is designed to employ the noise map to compute a noise reduction offset for the person feature embedding.

\textbf{Noise Extractor.} This module is composed of three standard 3×3 convolution layers containing a convolution, a ReLU, and a batch normalization, as shown in Figure \ref{fig:fmn}. The three convolution layers are obtained by removing all 1×1 convolution layers and residual structures in Layer4 of ResNet \cite{he2016deep}. In addition, to reduce the number of parameters, the number of channels in the middle layer is shrunk from 512 to 256.

The noise map output from the extractor is sequentially processed by a adaptive max pooling, a permutation, a learnable 2D-aware position encoding, and a layer normalization \cite{ba2016layer} in turn. To reduce the amount of computation, the adaptive maximum pooling is used to pool the noise map to one of size 7×7. We refer to the learnable 2D-aware position method in ViT \cite{dosovitskiy2020vit} to add a position encoding for each pixel in the noise map.

\textbf{Feature Denoiser.} Based on Transformer Decoder \cite{vaswani2017attention}, we design this denoiser, which takes a sequence of feature embeddings as the query and the noise maps corresponding to these embeddings as the key and value to infer the noise reduction offsets. As shown in Figure \ref{fig:fmn}, this module differs from Transformer Decoder in three aspects: \ding{172} We replace the self-attention module with a feed forward module. Because the self-attention causes the network to focus on a few members of the embedding sequence, and the members in this sequence are independent and equal in status, using the self-attention is inappropriate. \ding{173} The residual structure on the cross-attention module is removed. The residual structure allows the network to skip the cross-attention operation, which can lead to degradation of the denoiser. \ding{174} We use a batch normalization instead of a layer normalization to normalize the input of the feed forward module. Due to the equal status of the members in the sequence, the batch normalization performs better than the layer normalization. In addition, we design an Embedding Alignment (Emb Align) module to satisfy the input requirement of the cross-attention by adding some embeddings with values of 0 so that each noise map corresponds to an equal number of embeddings. Correspondingly, an Embedding De-alignment (Emb De-align) module is designed to remove the added embeddings.

\subsection{Bidirectional Online Instance Matching Loss}

OIM \cite{xiao2017joint} is the mainstream method for supervising the training of re-id subtask. OIM stores the representation proxies of all labeled persons in a look up table (LUT), $V=\left\{v_1, v_2, \cdots, v_L\right\}\in\mathbb{R}^{D\times L}$, consisting of $L$ vectors with $D$ dimension. After each iteration of training, the labeled feature representations of mini-batch is used to update the LUT with a momentum of $0.5$, while the unlabeled are saved by a circular queue (CQ), $U=\left\{u_1, u_2, \cdots, u_Q\right\}\in\mathbb{R}^{D\times Q}$. For a feature representation $x \in \mathbb{R}^D$ labeled as $l$, OIM computes the probability as follows:
\begin{equation}
    p_i = \frac{\exp{\left(v_i^\top x / \tau\right)}}{\sum_{j=1}^{L}{\exp{\left(v_j^\top x / \tau\right)}} + \sum_{k=1}^{Q}{\exp{\left(u_k^\top x / \tau\right)}}}
    \label{equ:oim-prob}
\end{equation}
where $\tau=1/30$ is hyper-parameter which adjusts the softness of the probability distribution. OIM objective is to minimize the expected negative log-likelihood:
\begin{equation}
    \mathcal{L}_{OIM} = -\mathrm{E}_{x}\left(\log{p_t}\right), t = 1, 2, \cdots, L
    \label{equ:oim}
\end{equation}

\begin{table*}[t]
    \scriptsize
    \centering
    \renewcommand{\arraystretch}{\tstretch}
    \begin{tabularx}{\linewidth}{
            >{\hsize=.3\hsize \linewidth=\hsize \raggedright\arraybackslash}X
            |>{\hsize=.21\hsize \linewidth=\hsize \raggedright\arraybackslash}X
            *{2}{|*{2}{>{\hsize=.09\hsize \linewidth=\hsize \centering\arraybackslash}X}}
        }
        \toprule
        \multicolumn{1}{c|}{} &
        \multicolumn{1}{c|}{} &
        \multicolumn{2}{c|}{\textbf{CUHK-SYSU}} &
        \multicolumn{2}{c}{\textbf{PRW}} \\
        \cline{3-6}
        \multicolumn{1}{c|}{\multirow{-2}{*}{\textbf{Method}}} &
        \multicolumn{1}{c|}{\multirow{-2}{*}{\textbf{Backbone}}} &
        \multicolumn{1}{c}{mAP} &
        \multicolumn{1}{c|}{top-1} &
        \multicolumn{1}{c}{mAP} &
        \multicolumn{1}{c}{top-1} \\
        \midrule
        \multicolumn{6}{c}{\textit{Two-step}} \\
        IDE \cite{zheng2017person}  & ResNet-50 & -    & -    & 20.5 & 48.3 \\
        RDLR \cite{han2019re} & ResNet-50 & 93.0 & 94.2 & 42.9 & 70.2 \\
        TCTS \cite{wang2020tcts} & ResNet-50 & 93.9 & 95.1 & 46.8 & 87.5 \\
        \midrule
        \multicolumn{6}{c}{\textit{End-to-end}} \\
        OIM \cite{xiao2017joint} & ResNet-50 & 75.5 & 78.7 & 21.3 & 49.4\\
        NAE+ \cite{chen2020norm} & ResNet-50 & 92.1 & 92.9 & 44.0 & 81.1\\
        AlignPS+ \cite{yan2021anchor} & ResNet-50 & 94.0 & 94.5 & 46.1 & 82.1\\
        SeqNet \cite{li2021sequential} & ResNet-50 & 93.8 & 94.6 & 46.7 & 83.4\\
        SeqNet+CBGM \cite{li2021sequential} & ResNet-50 & 94.8 & 95.7 & 47.6 & 87.6\\
        GLCNet \cite{zheng2021global} & ResNet-50 & 95.5 & 96.1 & 46.7 & 84.9 \\
        COAT \cite{yu2022cascade} & ResNet-50 & 94.2 & 94.7 & 53.3 & 87.4\\
        COAT+CBGM \cite{yu2022cascade} & ResNet-50 & 94.8 & 95.2 & 54.0 & 89.1\\
        PSTR \cite{cao2022pstr} & PVTv2-B2 & 95.2 & 96.2 & 56.5 & \underline{89.7}\\
        SeqNeXt \cite{jaffe2023gallery} & ConvNeXt-B & 96.1 & 96.5 & 57.6 & 89.5\\
        SeqNeXt+GFN \cite{jaffe2023gallery} & ConvNeXt-B & \underline{96.4} & 97.0 & 58.3 & \textbf{92.4}\\
        SeqNet(SOLIDER) \cite{chen2023beyond} & SOLIDER(Swin-S) & 95.5 & 95.8 & \underline{59.8} & 86.7\\
        \textbf{SEAS(ours)} & ResNet-50 & 96.2 & \underline{97.1} & 52.0 & 85.7 \\
        \textbf{SEAS(ours)} & ConvNeXt-B & \textbf{97.1} & \textbf{97.8} & \textbf{60.5} & 89.5 \\
        \bottomrule
    \end{tabularx}
    \caption{Comparison of mAP and top-1 accuracy with state-of-the-art two-step and end-to-end methods on the benchmark CUHK-SYSU and PRW datasets. The best and second best scores are shown in bold and underlined, respectively.}
    \label{tab:sota}
\end{table*}

Although OIM achieves satisfactory results, we still obvious its limitations. Since the probability is calculated by the softmax function in Equation \ref{equ:oim-prob}, OIM encourages learning of intra-person similarity while being insensitive to inter-person differences. This makes the feature representation lack differentiation, so we design a triplet loss to encourage learning about the gap between persons. Specifically, we set a hyper-parameter $M$ to control the minimum margin of the inter-person cosine similarity relative to the intra-person. In each mini-batch, $\mathcal{B} \in \mathbb{R}^{D \times B}$, given a feature representation $x \in \mathbb{R}^D$ labeled as $l$, we sample a vector set $X_1 \in \mathbb{R}^{D \times P}$ consisting of $P$ members with the same label $l$ and a vector set $X_0 \in \mathbb{R}^{D \times N}$ consisting of $N$ members with labels other than $l$ from the concatenated set, $\mathcal{B} \cup V$, of the mini-batch and the LUT. The triplet loss can be calculated as:
\begin{equation}
    \mathcal{L}_{M} = \max\left\{ M - \left[\min\left(X^\top_1 x\right) - \max\left(X^\top_0 x\right)\right], 0 \right\}
    \label{equ:triplet}
\end{equation}
where $X^\top_1 x$ and $X^\top_0 x$ both represent a set of cosine similarities between the feature representation and members of the vector set. Finally, the Bidirectional Online Instance Matching (BOIM) loss is obtained by adding two terms together:
\begin{equation}
    \mathcal{L}_{BOIM} = \mathcal{L}_{M} + \mathcal{L}_{OIM}
    \label{equ:boim}
\end{equation}

%% file: sections/4_experiments.tex
\section{Experiments}
\label{sec:experiments}

\subsection{Datasets and Evaluation Metric}

\quad\textbf{CUHK-SYSU} \cite{xiao2017joint} is a large-scale person search dataset that contains 18,184 images collected from handheld cameras, movies, and TV shows, resulting in significant scene diversity. It encompasses 8,432 unique person identities and 96,143 annotated bounding boxes. The dataset is split into a training set with 5,532 identities and 11,206 images, and a test set with 2,900 query persons and 6,978 gallery images. For each query, the dataset defines a gallery size ranging from 50 to 4,000, with a default gallery size of 100 images.

\textbf{PRW} \cite{zheng2017person} is composed of video frames captured by six fixed cameras on a university campus. It contains 11,816 scene images with 932 distinct person identities and 43,110 annotated bounding boxes. The training set consists of 932 identities with 5,704 images, and the test set contains 2,057 query persons and 6,112 scene images. For each query, the dataset uses all images in the test set except for the query as the gallery.

\textbf{Evaluation Metric.} Following the established setting in previous work, the mean Average Precision (mAP) and top-1 accuracy (top-1) are employed to evaluate the performance for person search.

\subsection{Implementation}

We conduct all experiments on a NVIDIA A800 GPU and implement our model with PyTorch \cite{paszke2019pytorch}. ConvNeXt \cite{liu2022convnet} pre-trained on Imagenet \cite{deng2009imagenet} is adopted as the backbone network. During training, we set the batch size to 5 for CUHK-SYSU and 8 for PRW using Automatic Mixed Precision (AMP). Adam is used to optimize our model for 20 epochs with an initial learning rate of 0.0001, which is warmed up during the first epoch and reduced by a factor of 10 at epochs 8 and 14. For Equation \ref{equ:triplet}, we initialize the hyper-parameter $M$ as 0.25 for CUHK-SYSU and 0.35 for PRW. For MGE, the probability of the Drop Path is set to 0.1.

\subsection{Comparison to the State-of-the-Arts}

In this section, we compare our method with the state-of-the-art algorithms, including both two-step methods and end-to-end methods, on the benchmark CUHK-SYSU and PRW datasets.

\textbf{Results on CUHK-SYSU.} Table \ref{tab:sota} shows the performance on the CUHK-SYSU test set with the gallery size of 100. our SEAS achieves the best 97.1\% mAP and the highest 97.8\% top-1 accuracy, and with the same backbone outperforms the best two-step method TCTS \cite{wang2020tcts} in terms of mAP and top-1 accuracy by 2.3\% and 2.0\%, respectively, despite it employs two separate models for detection and re-id. Among end-to-end methods, our method performs better than the first one-stage anchor-free framework AlignPS+ \cite{yan2021anchor}, the two-stage representative work SeqNet \cite{li2021sequential}, and the transformer-based methods COAT \cite{yu2022cascade} and PSTR \cite{cao2022pstr}. For the current state-of-the-art SeqNeXt+GFN \cite{jaffe2023gallery} using the scene-binding strategy, our framework outperforms it by 1.5\% and 1.8\% in mAP and top-1 accuracy, respectively, when both employ ResNet-50 as the backbone, and by 0.6\% and 0.8\% when both use ConvNeXt-B as the backbone. Notably, when using ResNet-50, SEAS can even outperform some of the approaches based on more advanced backbones, such as PSTR, SeqNeXt \cite{jaffe2023gallery} and SOLIDER \cite{chen2023beyond}.

\begin{figure}[t]
    \centering
    \includegraphics[scale=0.57]{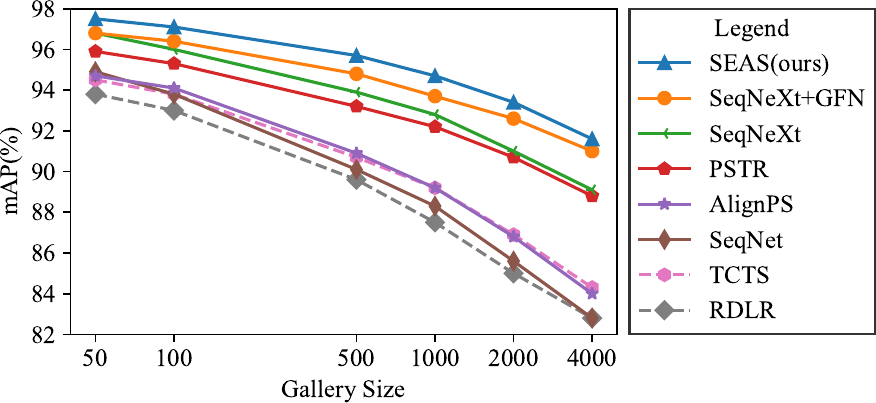}
    \caption{Comparison of mAP on CUHK-SYSU across various gallery sizes. The solid lines represent end-to-end methods and the dashed lines represent two-step methods.}
    \label{fig:cuhk}
\end{figure}

We further perform a comparison between our SEAS and state-of-the-art methods on the CUHK-SYSU test set with gallery sizes ranging from 50 to 4,000. Figure \ref{fig:cuhk} shows the performance curve of our method with some two-step and end-to-end methods in terms of mAP as the gallery size changes. Since it is a challenge for all compared methods to consider more distracting persons in the gallery set, the performance of them is reduced as the gallery size increases. However, our SEAS consistently outperforms all the two-step and end-to-end methods.

\begin{table}[t]
    \scriptsize
    \centering
    \renewcommand{\arraystretch}{\tstretch}
    \begin{tabularx}{\linewidth}{
            >{\hsize=.335\hsize\linewidth=\hsize\raggedright\arraybackslash}X
            |>{\hsize=.335\hsize\linewidth=\hsize\raggedright\arraybackslash}X
            |*{2}{>{\hsize=.165\hsize\linewidth=\hsize\centering\arraybackslash}X}
        }
        \toprule
        \multicolumn{1}{c|}{\textbf{Method}} &
        \multicolumn{1}{c|}{\textbf{Backbone}} &
        \multicolumn{1}{c}{\textbf{mAP}} &
        \multicolumn{1}{c}{\textbf{top-1}} \\
        \midrule
        SeqNeXt & ConvNeXt-B & 91.4 & 92.4 \\
        SeqNeXt+GFN & ConvNeXt-B & 92.0 & 93.1 \\
        \textbf{SEAS (ours)} & ConvNeXt-B & \textbf{94.0} & \textbf{95.2} \\
        \bottomrule
    \end{tabularx}
    \caption{Comparison of mAP and top-1 accuracy on the low-resolution subset of CUHK-SYSU.}
    \label{tab:cuhk}
\end{table}

Table \ref{tab:cuhk} shows the performance of our SEAS and the current best preforming method on the low-resolution subset of CUHK-SYSU, where the query image is low-resolution. Although the reduced resolution increases the difficulty of retrieval, our method with the same backbone outperforms SeqNeXt+GFN by 2.0\% and 2.1\% in mAP and top-1 accuracy, respectively, to 94.0\% and 95.2\%.

\textbf{Results on PRW.} Since PRW \cite{zheng2017person} has a smaller training set but a larger gallery size than CUHK-SYSU \cite{xiao2017joint}, all methods do not perform as well on this dataset as they do on CUHK-SYSU. SEAS achieves the highest score of 60.5\% on the key evaluation metric mAP, outperforming all two-step models by a wide margin. For the current best-performing method SeqNet(SOLIDER) \cite{chen2023beyond}, which replaces the backbone of SeqNet with Swin-Transformer whose pretraining weights are trained using a self-supervised training method called SOLIDER \cite{chen2023beyond}, our SEAS outperforms it by 0.7\% in mAP and 2.8\% in top-1 accuracy. Although SeqNeXt+GFN \cite{jaffe2023gallery} performs well in top-1 accuracy with its scene binding strategy, SEAS beats it by 2.2\% in mAP, a more important evaluation metric than top-1.

\begin{table}[t]
    \scriptsize
    \centering
    \renewcommand{\arraystretch}{\tstretch}
    \begin{tabularx}{\linewidth}{
            >{\hsize=.335\hsize\linewidth=\hsize\raggedright\arraybackslash}X
            |>{\hsize=.335\hsize\linewidth=\hsize\raggedright\arraybackslash}X
            |*{2}{>{\hsize=.165\hsize\linewidth=\hsize\centering\arraybackslash}X}
        }
        \toprule
        \multicolumn{1}{c|}{\textbf{Method}} &
        \multicolumn{1}{c|}{\textbf{Backbone}} &
        \multicolumn{1}{c}{\textbf{mAP}} &
        \multicolumn{1}{c}{\textbf{$\Delta$mAP}} \\
        \midrule
        HOIM & ResNet-50 & 36.5 & -21.8 \\
        NAE+ & ResNet-50 & 40.0 & -18.3 \\
        SeqNet & ResNet-50 & 43.6 & -14.7 \\
        SeqNet+CBGM & ResNet-50 & 44.3 & -14.0 \\
        AGWF & ResNet-50 & 48.0 & -10.3 \\
        COAT & ResNet-50 & 50.9 & -7.4 \\
        SeqNeXt & ConvNeXt-B & 55.3 & -3.0 \\
        SeqNeXt+GFN & ConvNeXt-B & 56.4 & -1.9 \\
        \textbf{SEAS (ours)} & ConvNeXt-B & \textbf{58.3} & - \\
        \bottomrule
    \end{tabularx}
    \caption{Performance of mAP on the PRW test set for query and gallery scene images from the different cameras.}
    \label{tab:prw}
\end{table}

We further compare the cross-camera performance of SEAS and state-of-the-art methods on the PRW test set. The gallery corresponding to each query in this comparison experiment is composed of all the images in the test set, excluding the images from the same camera as the query. As shown in Table \ref{tab:prw}, our method outperforms the existing methods with a clear margin. We attribute this to our scene-adaptive architecture, which generates scene-noise-free feature representations that effectively address the challenges posed by cross-camera or cross-scene retrieval.

\begin{figure}[t]
    \centering
    \includegraphics[scale=0.57]{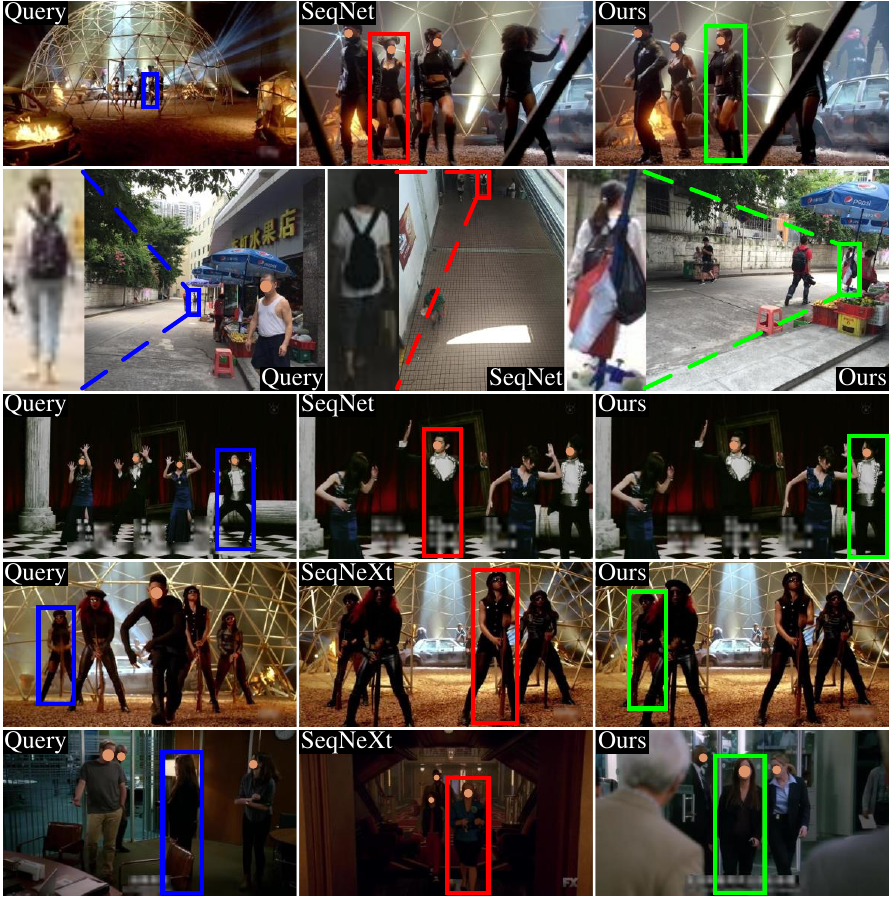}
    \caption{Qualitative comparison of SEAS with SeqNet and SeqNeXt on the CUHK-SYSU test set. The blue bounding boxes denote the queries, while the green and red bounding boxes denote correct and incorrect top-1 matches, respectively.}
    \label{fig:vis}
\end{figure}

\textbf{Qualitative Results.} Some example person search results are illustrated in Figure \ref{fig:vis}. We observe that our SEAS successfully handles cross-scene, low-resolution, occlusion, and viewpoint variation, while other state-of-the-art methods such as SeqNet \cite{li2021sequential} and SeqNeXt \cite{jaffe2023gallery} fail in these scenarios. This demonstrates the robustness of our methods.

\subsection{Ablation Study}

In this section, we perform analytical experiments to verify the effectiveness of each detailed component in our proposed framework.

\mdfdefinestyle{component}{
    linewidth=.6pt,
    innertopmargin=0.4em,
    innerbottommargin=0.4em,
    innerleftmargin=0.em,
    innerrightmargin=0.em,
    leftmargin=0.em,
    rightmargin=0.em,
}
\begin{figure}[t]
    \centering
    \begin{subfigure}{0.4\linewidth}
        \centering
        \begin{mdframed}[style=component]
            \centering
            \includegraphics[scale=0.57]{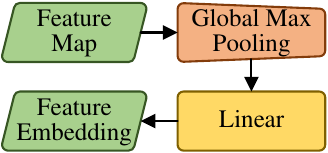}
        \end{mdframed}
        \caption{\scriptsize Global Feature Embedding}
        \label{fig:gfe}
    \end{subfigure}
    \hspace{-6pt}
    \begin{subfigure}{0.6\linewidth}
        \centering
        \begin{mdframed}[style=component]
            \centering
            \includegraphics[scale=0.57]{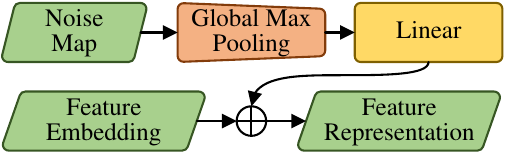}
        \end{mdframed}
        \caption{\scriptsize Linear Projection}
        \label{fig:fmn_linear}
    \end{subfigure}
    \caption{Details of the components used in the ablation study. (a) Details of the global feature embedding. (b) Details of the foreground modulation network with linear projection.}
\end{figure}

\begin{table}[t]
    \scriptsize
    \centering
    \renewcommand{\arraystretch}{\tstretch}
    \begin{tabularx}{\linewidth}{
            >{\hsize=0.25\hsize \linewidth=\hsize \raggedright\arraybackslash}X
            |*{2}{>{\hsize=.125\hsize \linewidth=\hsize \centering\arraybackslash}X}
            *{2}{|*{2}{>{\hsize=.125\hsize \linewidth=\hsize \centering\arraybackslash}X}}
        }
        \toprule
        \multicolumn{1}{c|}{} &
        \multicolumn{2}{c|}{\textbf{Feature Map}} &
        \multicolumn{2}{c|}{\textbf{CUHK-SYSU}} &
        \multicolumn{2}{c}{\textbf{PRW}} \\
        \cline{2-7}
        \multicolumn{1}{c|}{\multirow{-2}{*}{\textbf{Embedding}}} &
        \multicolumn{1}{c}{24×12} &
        \multicolumn{1}{c|}{12×6} &
        \multicolumn{1}{c}{mAP} &
        \multicolumn{1}{c|}{top-1} &
        \multicolumn{1}{c}{mAP} &
        \multicolumn{1}{c}{top-1} \\
        \midrule
        GFE & \ding{51} & \ding{51} & 94.5 & 94.8 & 43.3 & 79.4 \\
        MGE & \ding{51} & \ding{51} & \textbf{97.1} & \textbf{97.8} & 60.5 & \textbf{89.5} \\
        MGE & \ding{55} & \ding{51} & \textbf{97.1} & 97.6 & \textbf{61.0} & 88.3 \\
        \bottomrule
    \end{tabularx}
    \caption{Comparative results of the different feature embeddings.}
    \label{tab:emb}
\end{table}

\textbf{Multi-Granularity Embedding.} To validate the effectiveness of Multi-Granularity Embedding, we compare it with Global Feature Embedding (GFE) commonly used in other works. In GFE, as shown in Figure \ref{fig:gfe}, the global max pooling is used for the feature map to get a vector, which is then projected into a 512-dimensional embedding by the linear layer. As observed in Table \ref{tab:emb}, MGE outperforms GFE by a large margin, which indicates our strategy combining global and local information in different granularities is superior to a single global feature. Further, we evaluate the performance of our framework in the case of generating the feature embedding without using the 24×12 person feature map but only the 12×6 map that has been downsampled by Conv5. The results show that our strategy of encoding feature embeddings using feature maps from both levels has the best overall performance.

\begin{table}[b]
    \scriptsize
    \centering
    \renewcommand{\arraystretch}{\tstretch}
    \begin{tabularx}{\linewidth}{
            >{\hsize=0.3\hsize \linewidth=\hsize \raggedright\arraybackslash}X
            *{2}{|*{2}{>{\hsize=.15\hsize \linewidth=\hsize \centering\arraybackslash}X}}
        }
        \toprule
        \multicolumn{1}{c|}{\textbf{BNR loss}} &
        \multicolumn{1}{c}{\textbf{mAP}} &
        \multicolumn{1}{c|}{\textbf{top-1}} &
        \multicolumn{1}{c}{\textbf{$\Delta$mAP}} &
        \multicolumn{1}{c}{\textbf{$\Delta$top-1}} \\
        \midrule
        No BNR & 96.5 & 97.2 & - & - \\
        w/o BN & 96.8 & 97.4 & +0.3 & +0.2\\
        w/ BN & \textbf{97.1} & \textbf{97.8} & +0.6 & +0.6 \\
        \bottomrule
    \end{tabularx}
    \caption{Comparative results on CUHK-SYSU by employing different strategies for BNR loss.}
    \label{tab:bnr}
\end{table}

\textbf{Background Noise Reduction Loss.} To verify the effectiveness of our BNR design, we evaluate separately the performance of our framework in three scenarios: SEAS without BNR, BNR without BN, and BNR with BN, and report the results in Table \ref{tab:bnr}. We find that the BNR without and with BN both improve the mAP and top-1 accuracy. This indicates that our design enhances the network's ability to filter background noise. In addition, BNR without BN is to remove the BN term in Equation \ref{equ:bnr-prob}, at which time the output range of this equation changes from $ (0,1) $ to $ [0.5,1) $. The experimental results show that the absence of BN leads to a significant degradation in the performance of SEAS, which suggests that our use of BN to map the L$_2$-norm of feature embedding is necessary and effective.

\begin{table}[t]
    \scriptsize
    \centering
    \renewcommand{\arraystretch}{\tstretch}
    \begin{tabularx}{\linewidth}{
            >{\hsize=0.3\hsize \linewidth=\hsize \raggedright\arraybackslash}X
            *{2}{|*{2}{>{\hsize=.15\hsize \linewidth=\hsize \centering\arraybackslash}X}}
        }
        \toprule
        \multicolumn{1}{c|}{} &
        \multicolumn{2}{c|}{\textbf{CUHK-SYSU}} &
        \multicolumn{2}{c}{\textbf{PRW}} \\
        \cline{2-5}
        \multicolumn{1}{c|}{\multirow{-2}{*}{\textbf{FMN}}} &
        \multicolumn{1}{c}{mAP} &
        \multicolumn{1}{c|}{top-1} &
        \multicolumn{1}{c}{mAP} &
        \multicolumn{1}{c}{top-1} \\
        \midrule
        No FMN & 96.6 & 97.1 & 59.9 & 89.1 \\
        w/ Linear Projection & 96.7 & 97.4 & 58.7 & 88.5 \\
        w/ Cross-Attention & \textbf{97.1} & \textbf{97.8} & \textbf{60.5} & \textbf{89.5} \\
        \bottomrule
    \end{tabularx}
    \caption{Comparative results by employing different strategies for FMN.}
    \label{tab:fmn}
\end{table}

\textbf{Foreground Modulation Network.} Table \ref{tab:fmn} shows the performance of the FMN being employed with different strategies on the two benchmarks. First, to validate the effectiveness of the FMN, we removed the FMN and used the feature embedding generated by the BMN for person search. The results reveal that our strategy of employing scene information to eliminate foreground noise can significantly improve the retrieval accuracy. Furthermore, we compare the performance of using cross-attention in FMNs with linear projection, in which the noise map is sequentially processed through global maximum pooling and linear transformation to obtain the offsets used for correction, as shown in Figure \ref{fig:fmn_linear}. From Table \ref{tab:fmn}, we learn that the performance of cross-attention is ahead of the linear projection on both benchmarks, which indicates that the structure of the denoiser we designed can accurately calculate the noise reduction offset and thus effectively eliminate foreground noise.

\begin{table}[t]
    \scriptsize
    \centering
    \renewcommand{\arraystretch}{\tstretch}
    \begin{tabularx}{\linewidth}{
            >{\centering\arraybackslash}X
            *{2}{|*{2}{>{\centering\arraybackslash}X}}
        }
        \toprule
        \multicolumn{1}{c|}{} &
        \multicolumn{2}{c|}{\textbf{CUHK-SYSU}} &
        \multicolumn{2}{c}{\textbf{PRW}} \\
        \cline{2-5}
        \multicolumn{1}{c|}{\multirow{-2}{*}{\textbf{$M$}}} &
        \multicolumn{1}{c}{mAP} &
        \multicolumn{1}{c|}{top-1} &
        \multicolumn{1}{c}{mAP} &
        \multicolumn{1}{c}{top-1} \\
        \midrule
        0.20 & 97.0 & 97.4 & 59.6 & 88.3 \\
        0.25 & \textbf{97.1} & \textbf{97.8} & 59.8 & 88.4 \\
        0.30 & 96.7 & 97.4 & 60.1 & 87.0 \\
        0.35 & 96.7 & 97.4 & \textbf{60.5} & \textbf{89.5} \\
        0.40 & 96.5 & 97.0 & 60.2 & 89.1 \\
        \bottomrule
    \end{tabularx}
    \caption{Performance on the benchmark CUHK-SYSU and PRW datasets with different $M$ of BOIM.}
    \label{tab:margin}
\end{table}

\textbf{Analysis on Hyper-Parameter $M$.} We conduct experiments for the influence of hyper-parameter $M$ in Equation \ref{equ:triplet}. As shown in Table \ref{tab:margin}, when CUHK-SYSU is adopted as the experimental dataset, the best performances of both mAP and top-1 accuracy are achieved with $M=0.25$, while with $M=0.35$ for PRW. These results might be counterintuitive but validate the effectiveness of the hyper-parameter $M$ in controlling the margin of similarity between persons. Specifically, the training sets of CUHK-SYSU and PRW have 5,532 and 932 person identities, respectively. Since CUHK-SYSU has a higher number of identities, the margin between identities should be smaller. The experimental results verify our inference above, so setting the hyper-parameter $M$ is reasonable and effective.

%% file: sections/5_conclusion.tex
\section{Conclusion}
\label{sec:conclusion}

In this paper, we propose the Scene-Adaptive framework to achieve a consistent feature representation of the same person in diverse scenes. The inconsistency of the person features is due to the scene noise contained therein, which consists of background noise and foreground noise. For the two noises, we design a background modulation network and a foreground modulation network to achieve background noise filtering out and foreground noise elimination, respectively.